\pdfoutput=1

\documentclass[11pt]{article}

\usepackage{acl}
\usepackage{url}
\usepackage{hyperref}

\usepackage{times}
\usepackage{latexsym}
\usepackage[T1]{fontenc}
\usepackage[utf8]{inputenc}
\usepackage[nopatch=footnote]{microtype}
\usepackage{inconsolata}
\usepackage{amsmath}
\usepackage{amssymb}
\usepackage{graphicx}
\usepackage{booktabs}
\usepackage{float}
\usepackage{booktabs}
\usepackage{multirow}
\usepackage{array}
\usepackage{subfig}
\usepackage{amsmath}
\usepackage{amssymb}
\usepackage{bbm}

\usepackage{listings}
\title{Neurosymbolic Clinical Trial Matching \\ via LLM-Driven Abduction and Logical Verification}
\author{
  \textbf{Baiyang Qu}$^1$, \textbf{Leonardo Ranaldi}$^2$, \textbf{Xi Wang}$^3$, \textbf{Marco Valentino}$^3$ \\
  $^1$Department of Computer Science, University of Leicester, UK\\
  $^2$School of Informatics, University of Edinburgh, UK\\ 
  $^3$School of Computer Science, University of Sheffield, UK\\
  \texttt{bq19@leicester.ac.uk}\\
\texttt{leonardo.ranaldi@ed.ac.uk} \\
\texttt{\{xi.wang, m.valentino\}@sheffield.ac.uk}
}

\date{}

\begin{document}
\maketitle

\begin{abstract}
Large Language Models (LLMs) offer a promising path to automate Clinical Trial Matching (CTM), but still struggle with the deterministic verification required for complex eligibility criteria. Conversely, purely symbolic methods provide formal rigour but break down when faced with incomplete patient records and noisy clinical evidence. To bridge this gap, we investigate a hybrid framework for CTM combining LLMs with logical verification. In particular, we introduce an \emph{abductive neurosymbolic CTM framework} ($\alpha$NeSy-CTM), which leverages the linguistic and world knowledge in LLMs to support reasoning over noisy and underspecified clinical text.
Extensive evaluation demonstrates that $\alpha$NeSy-CTM substantially outperforms standalone LLM baselines, achieving up to 30\% relative improvement over zero-shot baselines. In addition, our analyses confirm the impact of abductive reasoning on CTM, with $\alpha$NeSy-CTM exhibiting improved accuracy, specificity, and robustness over a non-abductive neurosymbolic setting. Furthermore, $\alpha$NeSy-CTM and Chain-of-Thought (CoT) reasoning prove highly complementary, highlighting the potential for a hybrid routing policy.
Ultimately, this paper demonstrates the impact of neurosymbolic methods for automating CTM, providing a path toward the next generation of auditable, LLM-driven clinical applications.

\end{abstract}

\section{Introduction}
\label{sec:intro}

Clinical trials serve as the fundamental bridge between biomedical discovery and life-saving treatments, yet the translation of research into practice is frequently stalled by severe recruitment bottlenecks~\citep{treweek2013strategies}. As trial protocols grow in volume and complexity, manual screening of patient records against eligibility criteria has become an unsustainable, resource-heavy process prone to human error. Recently, Large Language Models (LLMs) have emerged as a promising avenue for automating Clinical Trial Matching (CTM)~\citep{jin2024trialgpt,wornow2024zeroshotctm,rybinski2023questionnaires,ghosh2025llmctr,wong2023scaling}. Given their ability to handle linguistic variation across clinical narratives, LLMs offer a promising way to map unstructured patient records onto structured requirements~\cite{klusty2025leveragingllmsstructureddata}.

However, the profound representational gap between natural language and formal eligibility logic remains a critical barrier~\cite{fernando-lopez-ponce-bel-enguix-2025-limits}. CTM fundamentally requires deterministic verification of strict numerical thresholds, temporal windows, and nested logical dependencies~\cite{roberts2022overview}. In these high-stakes settings, LLMs remain highly unreliable; their propensity for lack of verifiable reasoning traces preclude safe clinical deployment~\citep{rudin2019stop,jullien2024safebiomedical}. Conversely, symbolic methods and rule-based systems excel at logical rigour but are notoriously brittle when confronted with the fragmented and incomplete information typical of clinical records~\cite{CDSS}.

\begin{figure*}[t]
\centering
\includegraphics[width=\textwidth]{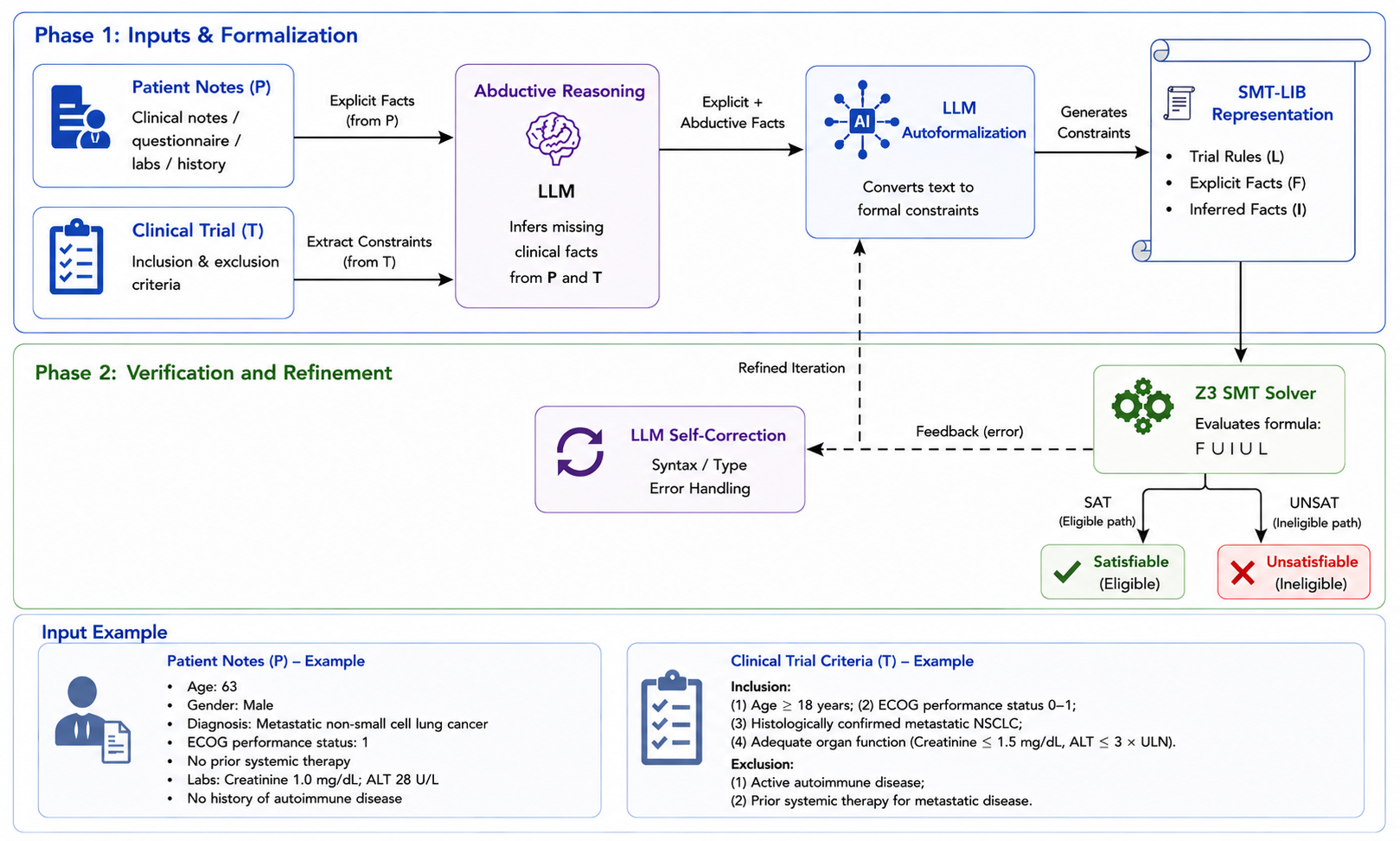}
\caption{Abductive Neurosymbolic Clinical Trial Matching. The framework consists of four main stages: abductive reasoning, autoformalization, logical verification, and refinement. LLM-driven abduction and formalisation handle incomplete and noisy information prevalent in clinical text, while Z3 determines eligibility via satisfiability checks.}
\label{fig:pipeline}
\end{figure*}

To bridge this gap, this paper investigates the transition of LLM-assisted CTM toward a neurosymbolic framework \cite{xu-etal-2026-adaptive,quan2024explanationrefiner,ye2023satlm,pan2023logic,xu2024symboliccot,ranaldi2025quasisymbolic}. In particular, we examine a hybrid framework that frames trial-patient matching as a constraint satisfaction problem, utilising LLMs to autoformalise patient profiles and eligibility criteria into symbolic constraints \cite{xu-etal-2026-adaptive,ye2023satlm}. 
These constraints are then evaluated by Z3 \cite{de2008z3}, a Satisfiability Modulo Theories (SMT) solver that acts as a deterministic logical engine.
To address the pervasive challenge of linguistic noise and underspecified clinical evidence, we propose to ground neuro-symbolic CTM in abductive reasoning. Specifically, we introduce $\alpha$NeSy-CTM, a framework that leverages LLMs for abduction to generate explicit, minimal evidence sets from implied clinical and linguistic cues, ensuring that information about eligibility and patient profiles is fully specified for robust formalization (Figure~\ref{fig:pipeline}).

To evaluate the effectiveness of our framework, we conduct experiments across the TREC Clinical Trials 2021--2023 benchmarks \cite{roberts2022overview,rybinski2023questionnaires}. The corresponding experimental results show that integrating $\alpha$NeSy-CTM yields consistent, multifaceted improvements over  LLM baselines, particularly as criteria complexity increases. On the TREC 2023 dataset, $\alpha$NeSy-CTM achieves a 75.71\% average accuracy, surging past the 58.17\% Zero-shot baseline. Crucially, our experiments confirm the impact of abductive reasoning on robustness, with $\alpha$NeSy-CTM suffering only a 2\% degradation on controlled evidence perturbations compared to an 8.5\% drop in a non-abductive neurosymbolic counterpart. Additionally, we found a complementarity between neurosymbolic methods and Chain-of-Thought (CoT). In particular, $\alpha$NeSy-CTM recovers 18\% of cases where CoT fails, while CoT succeeds on 12\% of cases where $\alpha$NeSy-CTM fails.

Ultimately, this work demonstrates the potential of LLM-driven abductive neurosymbolic reasoning for automating CTM, paving the way for the next generation of auditable clinical AI applications. 


\section{Methodology}
\label{sec:method}

Given a patient description $P$ and a clinical trial report $T$, both expressed in natural language, Clinical Trial Matching (CTM) aims to determine patient eligibility ($y \in \{0,1\}$). This task is challenging because clinical text is unstructured, trial criteria contain strict formal constraints, and patient evidence is frequently underspecified. 

\subsection{Problem Formalization}
\label{sec:formalization}

For a given patient--trial pair, CTM defines a variable space $V = \{v_1, \ldots, v_k\}$ covering criterion-relevant attributes such as age, prior treatments, diagnoses, and temporal indicators. Unlike pure boolean variables, medical constraints require typed domains; thus, each variable $v_i \in V$ is associated with a specific domain of possible values (e.g., integers, reals, booleans) evaluated under a background theory $\mathcal{T}$ (such as linear arithmetic).

The eligibility criteria in a clinical trial report $T$ are mapped into a logical theory via an extraction function $\phi$:
\begin{equation}
    L = \phi(T)
\end{equation}
where $L$ is represented as a set of symbolic constraints over $V$. In parallel, the patient profile $P$ defines a fact set via a mapping function $\psi$:
\begin{equation}
    F = \psi(P)
\end{equation}
where $F$ consists of ground truths, assigning specific values to a subset of the variables in $V$.

Eligibility is then cast as a Satisfiability Modulo Theories (SMT) problem \cite{xu-etal-2026-adaptive}. A patient is predicted to be eligible ($y=1$) if the combined theory formed by the observed facts and the trial eligibility constraints admits at least one valid model under the background theory $\mathcal{T}$. If any contradiction is identified, the patient is regarded as not eligible ($y=0$).

Formally, the eligibility prediction $y \in \{0, 1\}$ is determined by evaluating the $\mathcal{T}$-satisfiability of the union of these sets. In particular, the binary prediction $y$ is based on whether $F \cup L$ belongs to the set of satisfiable SMT formulas:
\begin{equation}
    y = \begin{cases} 
        1 & \text{if } F \cup L \in \text{SMT}(\mathcal{T}) \\ 
        0 & \text{otherwise} 
    \end{cases}
\end{equation}
Alternatively, expressed through logical entailment, where $\bot$ represents a contradiction and $\models_{\mathcal{T}}$ denotes entailment modulo theory $\mathcal{T}$:
\begin{equation}
    y = \begin{cases} 
        1 & \text{if } F \cup L \not\models_{\mathcal{T}} \bot \\ 
        0 & \text{if } F \cup L \models_{\mathcal{T}} \bot 
    \end{cases}
\end{equation}

\subsection{Neurosymbolic CTM (NeSy-CTM)}
\label{sec:nesy_ctm}

In a standard neurosymbolic approach, the formalization is implemented by prompting an LLM to derive formal constraints ($L$) and logical facts ($F$) from input text, while delegating satisfiability verification to the Z3 solver \citep{de2008z3, xu-etal-2026-adaptive,ye2023satlm}. The pipeline consists of three stages:

\begin{enumerate}
    \item \textbf{Autoformalization:} An LLM translates eligibility criteria into SMT-LIB constraints (e.g., \texttt{assert >= Age 18}). Information in the patient profile is used to derive additional assertions by extracting values for the variables in the constraints (e.g., \texttt{assert Age = 20}).
    \item \textbf{Symbolic Verification:} The formalized facts $F$ are checked against the criteria $L$ using Z3, yielding a deterministic \texttt{sat} (eligible) or \texttt{unsat} (ineligible) decision.
    \item \textbf{Self-Correction Loop:} If Z3 reports syntax or type errors, the LLM receives error feedback for iterative refinement, acting as tool-augmented reasoning \citep{schick2023toolformer}.
\end{enumerate}

\paragraph{Example.} Suppose the trial requires \texttt{Age >= 18}, while the patient note states that the patient is 52 years old. The autoformalizer produces:
\begin{quote}
    \texttt{(declare-const Age Int)}\\
    \texttt{(assert (= Age 52))}\\
    \texttt{(assert (>= Age 18))}\\
    \texttt{(check-sat)}
\end{quote}
In this case, Z3 will return \texttt{sat} since the assertions do not produce any logical contradiction under integer arithmetic.

\subsection{Abductive Reasoning for CTM ($\alpha$NeSy-CTM)}
\label{sec:abductive}

The standard setting is effective when patient profiles and eligibility criteria are fully specified and perfectly aligned. However, in real-world scenarios, linguistic noise (e.g., lexical variations) and implied clinical assumptions can prevent robust satisfiability verification. For the standard setting, if the set of variables in $L$ does not exactly match those in $F$, the entire formalization will be underspecified. This can result in Z3 overestimating eligibility. To mitigate this challenge, $\alpha$NeSy-CTM introduces an additional abductive reasoning stage where the LLM fills knowledge and lexical gaps for a more reliable formalization.

The abductive extension proceeds as follows:
\begin{enumerate}
    \item \textbf{Abductive Reasoning:} Given the patient profile and eligibility criteria, the LLM generates a conservative set of inferred facts $I$ that are not explicitly stated but are supported by the text. These may include matched acronyms, treatment history, and disease-state indicators. For example, phrases such as ``walks independently'' and ``performs daily activities without limitation'' can indicate a low ECOG status, without assigning an overly specific value unless it is explicitly supported by the text.
    \item \textbf{Augmented Verification:} The abductive output is merged with the observed patient facts and the constraints to form the final theory $F \cup I \cup L$. The $\mathcal{T}$-satisfiability of the theory is then checked by Z3.
\end{enumerate}

\paragraph{Example.} Consider the patient note: \textit{``A 63-year-old man who walks independently, is able to shop for groceries, and reports no assistance with daily activities.''} A trial requires \texttt{ECOG >= 3}, which implies a patient capable of limited self-care, confined to a bed or chair more than 50\% of waking hours. However, from the patient description, the abductive reasoning stage can infer a low ECOG status, such as \texttt{ECOG = 0} or \texttt{ECOG = 1}, and add it to the theory. These new facts inferred via abduction are crucial to fully determining ineligibility. In a standard neurosymbolic approach, the ECOG value could remain unspecified, preventing Z3 from reliably identifying the contradiction.

\section{Empirical Evaluation}
\label{sec:experiments}

\subsection{Experimental Setup}

\paragraph{Dataset.} We evaluate binary eligibility prediction (\textit{Eligible} vs. \textit{Ineligible}) on the TREC Clinical Trials Track (2021--2023), a set of benchmarks for clinical trial matching with different patient representations and evidence structures~\citep{roberts2022overview,rybinski2023questionnaires}. Additional details on dataset statistics and preprocessing steps are included in Appendix~\ref{app:dataset_construction}.

\paragraph{TREC 2021.}
The 2021 edition established the foundational matching task with 75 synthetic patient cases against clinical trial descriptions derived from ClinicalTrials.gov \cite{roberts2022overview}. 

\paragraph{TREC 2022.}
This edition uses 50 topics and places greater emphasis on numerically salient evidence, including list-structured laboratory values and criterion formulations that more often hinge on threshold comparisons and treatment-history constraints~\citep{roberts2022overview}.

\paragraph{TREC 2023.}
This edition shifts patient representation toward semi-structured questionnaires, making fragmented or underspecified evidence more central to the matching problem and increasing the importance of reasoning under missing and fragmented information~\citep{rybinski2023questionnaires}.

\paragraph{Models.}
We compare \textbf{$\alpha$NeSy-CTM} against several baselines. In particular, we also implement a non-abductive neurosymbolic setting following a standard three-stage autoformalization-verification-refinement pipeline (NeSy-CTM), and compare $\alpha$NeSy-CTM against LLM-only baselines, including \textbf{Zero-shot} \cite{kojima2022zeroshot} (direct binary classification), \textbf{CoT} \cite{DBLP:journals/corr/abs-2201-11903} (binary classification supported by step-by-step natural language reasoning), and \textbf{QuaSAR} \cite{ranaldi2025quasisymbolic} (binary classification supported by step-by-step natural language reasoning augmented with symbolic formalisation). 
Experiments utilize three LLM backbones ($T=0.1$): \texttt{gpt-oss-120b}, \texttt{olmo-3-32b-think}, \texttt{qwen3-vl-32b-instruct}~\citep{openai2025gptoss,olmo2025olmo3,bai2025qwen3vl}. The symbolic engine is Z3 v4.12.2. The prompts adopted for the experiments are reported in Appendix~\ref{sec:appendix_prompts}.

\begin{table*}[t]
\small
\centering
\setlength{\tabcolsep}{4pt}
\renewcommand{\arraystretch}{1.06}
\resizebox{\textwidth}{!}{%
\begin{tabular}{ll|ccccc|ccccc|ccccc}
\toprule
Model & Method & \multicolumn{5}{c}{TREC 2021} & \multicolumn{5}{c}{TREC 2022} & \multicolumn{5}{c}{TREC 2023} \\
\cmidrule(lr){3-7}\cmidrule(lr){8-12}\cmidrule(lr){13-17}
 & & Acc & Sens & Spec & Macro-F1 & MCC & Acc & Sens & Spec & Macro-F1 & MCC & Acc & Sens & Spec & Macro-F1 & MCC \\
\midrule
\multirow{5}{*}{gpt-oss-120b} & Zero-shot & 68.0 & 85.0 & 51.0 & 67.0 & 38.3 & 73.0 & \textbf{90.0} & 56.0 & 72.2 & 48.9 & 59.5 & 94.0 & 25.0 & 54.0 & 26.3 \\
 & CoT & 71.5 & \textbf{87.0} & 56.0 & 70.8 & \textbf{45.2} & 73.5 & 88.0 & 59.0 & 72.9 & 49.1 & 62.0 & \textbf{96.0} & 28.0 & 57.0 & 32.7 \\
 & QuaSAR & 71.5 & 74.0 & 69.0 & 71.5 & 43.1 & 71.0 & 69.0 & 73.0 & 71.0 & 42.0 & 70.5 & 84.0 & 57.0 & 70.0 & 42.6 \\
 & NeSy-CTM & \textbf{72.0} & 62.0 & 82.0 & \textbf{71.7} & 44.9 & 75.0 & 60.0 & \textbf{90.0} & 74.4 & 52.4 & 74.5 & 79.0 & 70.0 & 74.5 & 49.2 \\
 & $\alpha$NeSy-CTM & 70.5 & 58.0 & \textbf{83.0} & 70.0 & 42.3 & \textbf{76.5} & 64.0 & 89.0 & \textbf{76.1} & \textbf{54.7} & \textbf{76.0} & 77.0 & \textbf{75.0} & \textbf{76.0} & \textbf{52.0} \\
\midrule
\multirow{5}{*}{olmo-3-32b-think} & Zero-shot & 64.5 & \textbf{84.0} & 45.0 & 63.1 & 31.5 & 60.5 & \textbf{82.0} & 39.0 & 58.6 & 23.3 & 55.0 & 84.0 & 26.0 & 50.9 & 12.3 \\
 & CoT & 62.8 & 68.0 & 57.6 & 62.7 & 25.7 & 65.8 & 67.7 & 63.9 & 65.8 & 31.6 & 55.5 & 57.0 & 54.0 & 55.5 & 11.0 \\
 & QuaSAR & 54.5 & 78.0 & 31.0 & 51.8 & 10.2 & 61.0 & 75.0 & 47.0 & 60.2 & 22.9 & 61.5 & \textbf{90.0} & 33.0 & 58.1 & 28.0 \\
 & NeSy-CTM & \textbf{70.5} & 53.0 & \textbf{88.0} & \textbf{69.6} & \textbf{43.8} & 70.0 & 51.0 & 89.0 & 68.9 & 43.2 & 76.0 & 74.0 & 78.0 & 76.0 & 52.0 \\
 & $\alpha$NeSy-CTM & 70.3 & 53.0 & 87.9 & 69.5 & 43.6 & \textbf{71.5} & 48.0 & \textbf{95.0} & \textbf{69.8} & \textbf{48.7} & \textbf{76.5} & 71.0 & \textbf{82.0} & \textbf{76.4} & \textbf{53.3} \\
\midrule
\multirow{5}{*}{qwen3-vl-32b-instruct} & Zero-shot & 61.0 & \textbf{91.0} & 31.0 & 57.1 & 27.5 & 61.5 & \textbf{94.0} & 29.0 & 57.0 & 30.3 & 59.5 & \textbf{96.0} & 23.0 & 53.3 & 27.8 \\
 & CoT & 64.5 & 84.0 & 45.0 & 63.1 & 31.5 & 67.0 & 90.0 & 44.0 & 65.2 & 38.3 & 63.0 & 94.0 & 32.0 & 59.1 & 33.1 \\
 & QuaSAR & 65.5 & 72.0 & 59.0 & 65.3 & 31.3 & 65.0 & 66.0 & 64.0 & 65.0 & 30.0 & 69.5 & 90.0 & 49.0 & 68.2 & 42.8 \\
 & NeSy-CTM & \textbf{68.5} & 65.0 & 72.0 & \textbf{68.5} & \textbf{37.1} & \textbf{73.0} & 66.0 & 80.0 & \textbf{72.9} & 46.5 & 72.5 & 85.0 & 60.0 & 72.1 & 46.5 \\
 & $\alpha$NeSy-CTM & 66.5 & 58.0 & \textbf{75.0} & 66.3 & 33.5 & \textbf{73.0} & 60.0 & \textbf{86.0} & 72.5 & \textbf{47.6} & \textbf{74.6} & 79.0 & \textbf{70.3} & \textbf{74.6} & \textbf{49.5} \\
\midrule
\multirow{5}{*}{Average} & Zero-shot & 64.5 & \textbf{86.7} & 42.3 & 62.7 & 32.4 & 65.0 & \textbf{88.7} & 41.3 & 62.9 & 34.1 & 58.0 & \textbf{91.3} & 24.7 & 52.8 & 21.5 \\
 & CoT & 66.3 & 79.7 & 52.8 & 65.7 & 33.8 & 68.8 & 81.9 & 55.5 & 68.2 & 38.9 & 60.2 & 82.3 & 38.0 & 58.1 & 22.7 \\
 & QuaSAR & 63.8 & 74.7 & 53.0 & 63.4 & 28.3 & 65.7 & 70.0 & 61.3 & 65.6 & 31.5 & 67.2 & 88.0 & 46.3 & 65.7 & 37.8 \\
 & NeSy-CTM & \textbf{70.3} & 60.0 & 80.7 & \textbf{70.0} & \textbf{41.6} & 72.7 & 59.0 & 86.3 & 72.2 & 47.1 & 74.3 & 79.3 & 69.3 & 74.3 & 48.9 \\
 & $\alpha$NeSy-CTM & 69.1 & 56.3 & \textbf{81.9} & 68.6 & 39.6 & \textbf{73.7} & 57.3 & \textbf{90.0} & \textbf{73.0} & \textbf{50.1} & \textbf{75.7} & 75.7 & \textbf{75.7} & \textbf{75.7} & \textbf{51.4} \\
\bottomrule
\end{tabular}
}
\caption{Overall results under model-specific alignment. Each year reports Accuracy (\%), Sensitivity (\%), Specificity (\%), Macro-F1 (\%), and MCC. Best accuracy values are in bold.}
\label{tab:main_res}
\end{table*}

\begin{figure}[t]
\centering
\includegraphics[width=\linewidth]{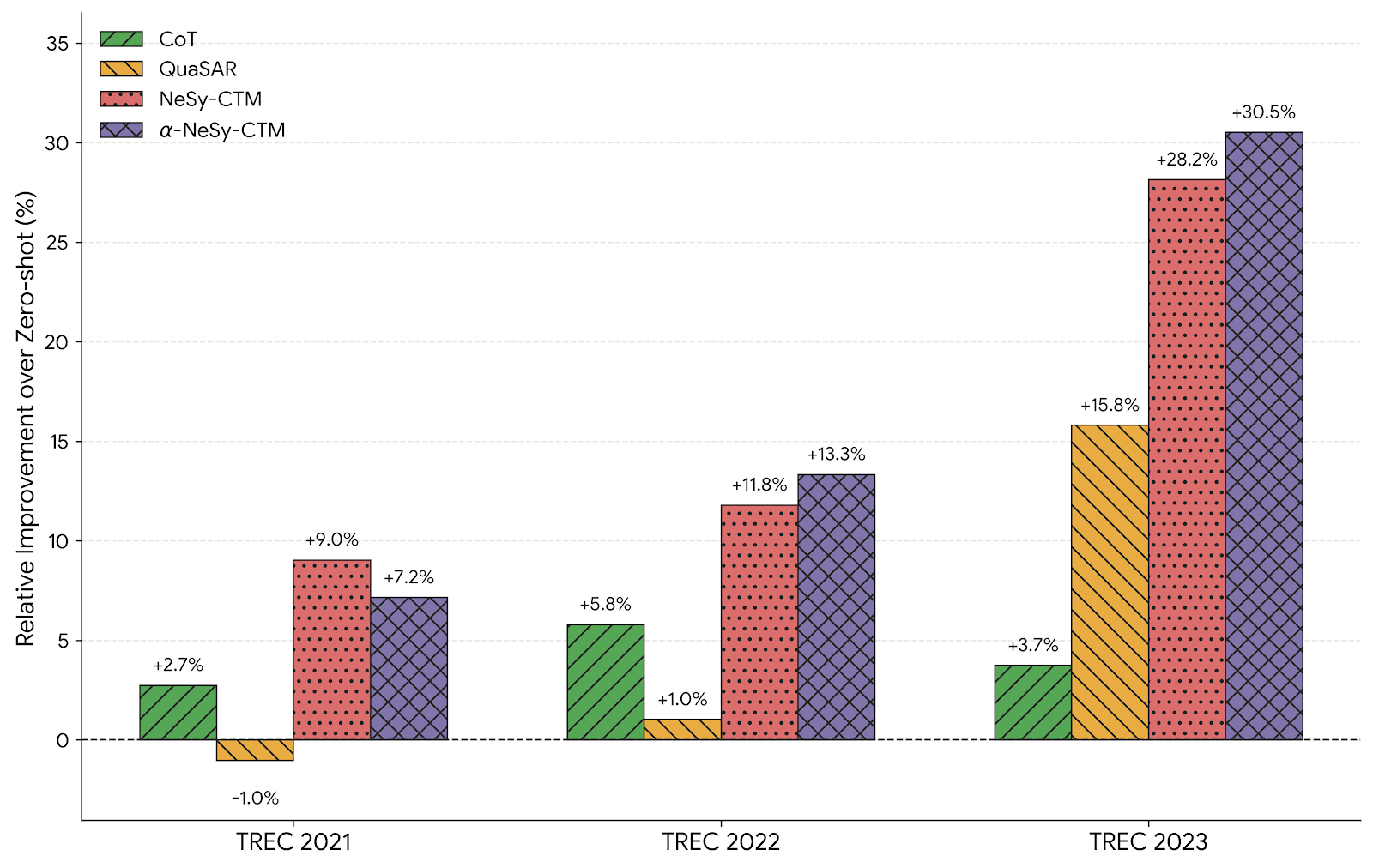}
\caption{Relative improvement over Zero-shot (\%) using the 3-backbone average from Table~\ref{tab:main_res}. Improvements are modest in 2021--2022, then surge in 2023, with $\alpha$NeSy-CTM showing the strongest uplift.}
\label{fig:table2_key_messages}
\end{figure}

\subsection{Results}
\label{sec:results}

The main results are reported in Table \ref{tab:main_res}. We compare the different methods using Accuracy (Acc), Sensitivity (Sens), Specificity (Spec), Macro-F1, and Matthews Correlation Coefficient (MCC).

\paragraph{Neurosymbolic methods shift the sensitivity-specificity trade-off toward balanced accuracy.} 
Table~\ref{tab:main_res} shows that the neurosymbolic variants consistently emerge as the strongest methods, with $\alpha$NeSy-CTM achieving the best average accuracy on the 2022 and 2023 splits, closely followed by NeSy-CTM. Furthermore, as illustrated in Figure~\ref{fig:table2_key_messages}, the relative improvement over the Zero-shot baseline surges dramatically on the challenging TREC 2023 dataset, with $\alpha$NeSy-CTM showing the strongest overall uplift. These results demonstrate not merely an absolute gain in accuracy, but a fundamental shift in the operational trade-off between sensitivity and specificity. Standalone LLM baselines achieve high sensitivity, particularly on TREC 2023, but do so at the cost of drastically over-predicting eligibility. For instance, the average Zero-shot system reaches 91.34\% sensitivity while severely compromising specificity (24.66\%), a permissive failure mode mirrored by CoT (38.00\% specificity). By contrast, $\alpha$NeSy-CTM maintains a substantially more balanced operational profile on TREC 2023, achieving 75.66\% sensitivity, 75.74\% specificity, 75.70 Macro-F1, and the highest MCC (0.514). 

\paragraph{Stricter logical boundaries drastically reduce false positives.} 
This dramatic improvement in specificity is critical for real-world clinical trial matching, where eligible patients are typically rare and false-positive reviews dominate the human workload. To illustrate the operational consequences under a hypothetical low-prevalence setting, consider screening 1,000 patients for a trial where only 5\% are truly eligible. Operating at the average TREC 2023 Zero-shot performance, the system would flag roughly 762 patients for manual review, generating about 716 false positives. Under the $\alpha$NeSy-CTM operating point, the review list shrinks to approximately 268 patients with around 230 false positives. While this stricter logical boundary misses slightly more eligible cases than the highly permissive LLM baselines, it produces a drastically more usable shortlist for clinical coordinators, directly combating alert fatigue.

\paragraph{Formal verification provides maximal gains on challenging, underspecified datasets.} 
In Table~\ref{tab:main_res}, TREC 2023 is the most challenging split for LLM-only inference: Zero-shot reaches its lowest accuracy (58.00\%) and specificity (24.66\%), while CoT also remains weak in specificity (38.00\%). In contrast, the solver-backed variants remain above 74\% accuracy, and $\alpha$NeSy-CTM achieves the best 2023 MCC (0.514). The larger separation on this split suggests that formal verification and controlled abduction are especially valuable when prompt-only inference becomes overly permissive, without requiring a stronger claim about dataset-level missingness.

\paragraph{Balanced classification creates a more reliable filter for human-in-the-loop screening.} 
Finally, the MCC confirms that neurosymbolic methods act as highly balanced classifiers, with a substantially lower false-positive bias than permissive candidate-generation baselines. On TREC 2023, MCC jumps from 0.215 (Zero-shot) and 0.227 (CoT) to 0.489 (NeSy-CTM) and 0.514 ($\alpha$NeSy-CTM). This indicates a stronger correlation with gold standard labels across both positive and negative classes. These results demonstrate that symbolic reasoning safely reduces the false-positive burden while retaining competitive recall. Ultimately, $\alpha$NeSy-CTM offers the most practical real-world trade-off for human-in-the-loop pre-screening.

\begin{table}[t]
\centering
\scriptsize
\setlength{\tabcolsep}{2pt}
\renewcommand{\arraystretch}{1.08}
\resizebox{\columnwidth}{!}{%
\begin{tabular}{@{}lccc@{\hspace{6pt}}cccc@{}}
\toprule
\textbf{Variant} & \textbf{F} & \textbf{S} & \textbf{A} & \textbf{2021} & \textbf{2022} & \textbf{2023} & \textbf{Avg} \\
\midrule
$\alpha$-NeSy & $\checkmark$ & $\checkmark$ & $\checkmark$ & 69.12 & \textbf{73.67} & \textbf{75.71} & \textbf{72.83} \\
NeSy & $\checkmark$ & $\checkmark$ & $\times$ & \textbf{70.33} & 72.67 & 74.33 & 72.44 \\
Form.-only$^{\dagger}$ & $\checkmark$ & $\times$ & $\times$ & 63.83 & 65.67 & 67.17 & 65.56 \\
LLM-direct$^{\ddagger}$ & $\times$ & $\times$ & $\times$ & 64.50 & 65.00 & 58.00 & 62.50 \\
\bottomrule
\end{tabular}}
\caption{Combined ablation configuration and results (\%) using the 3-backbone average values from Table~\ref{tab:main_res}. Configuration columns denote formalization (F), solver (S), and abduction (A). $\alpha$-NeSy = $\alpha$NeSy-CTM; NeSy = NeSy-CTM; $^{\dagger}$ uses a quasi-symbolic protocol (structured formalization without solver or abduction); $^{\ddagger}$ corresponds to the Zero-shot baseline.}
\label{tab:ablation_combined}
\end{table}

\begin{figure*}[t]
\centering
\includegraphics[width=\textwidth]{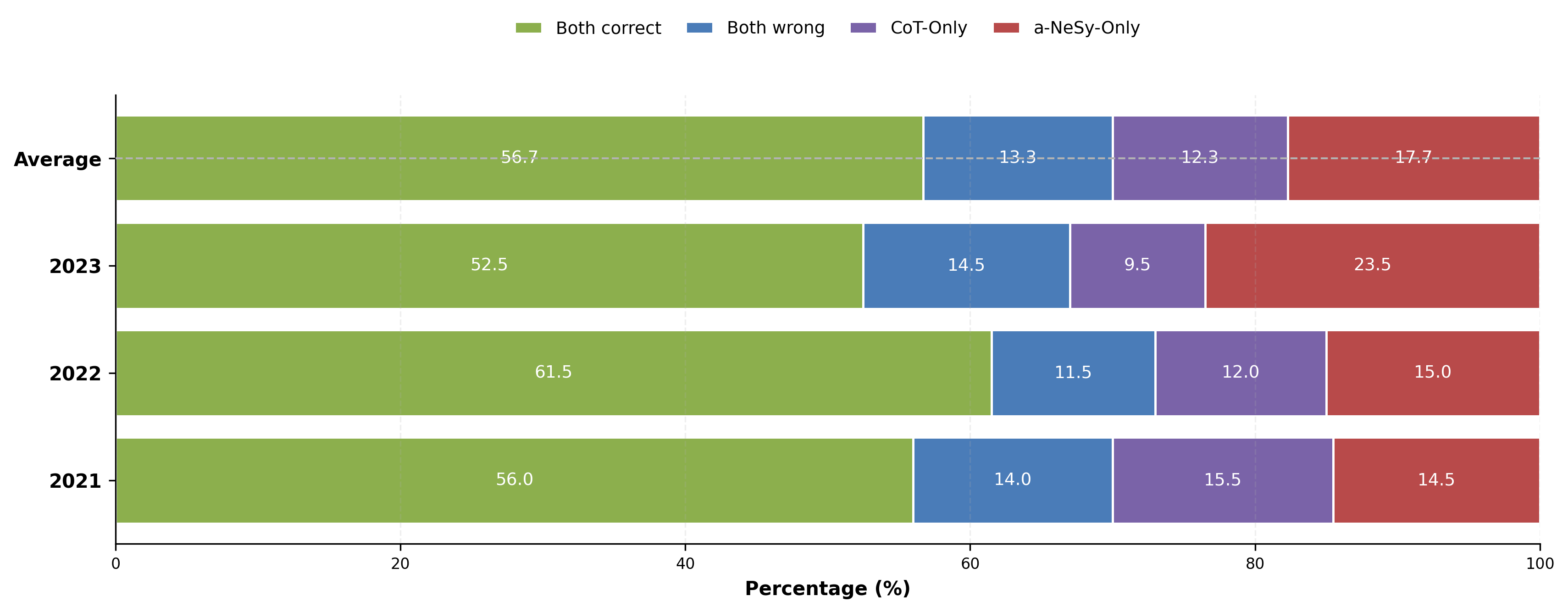}
\caption{Stacked intersection view for $\alpha$NeSy-CTM vs. CoT on the 600 paired samples from TREC 2021--2023. Each bar decomposes the yearly outcome distribution into Both correct, Both wrong, CoT-Correct, and $\alpha$NeSy-CTM-Correct; exact counts are reported in Appendix Table~\ref{tab:intersection}.}
\label{fig:intersection_stacked}
\end{figure*}

\subsection{Ablation Study}

To systematically evaluate the contribution of each module in our neurosymbolic architecture, we define four ablation variants that progressively integrate structured formalisation (F), solver-based verification (S), and abductive reasoning (A). Table~\ref{tab:ablation_combined} outlines these configurations and reports their 3-backbone average accuracies.

The results reveal a step-wise performance improvement. Simply translating criteria into a structured formalization without a solver (Form.-only) yields only marginal gains over the Zero-shot baseline (LLM-direct). However, the introduction of the Z3 solver (NeSy-CTM) results in a substantial performance leap across all years, underscoring the necessity of deterministic verification over raw LLM reasoning. Finally, integrating the abductive layer ($\alpha$NeSy-CTM) pushes the framework to its highest average performance (72.83\%).

The ablation results further clarify the role of each module. NeSy-CTM slightly outperforms $\alpha$NeSy-CTM in 2021 (70.33\% vs. 69.12\%), suggesting that when direct verification is sufficient, additional assumptions are not always beneficial. In 2022--2023, however, $\alpha$NeSy-CTM achieves the strongest performance, and in 2023 it improves over NeSy-CTM in both specificity (75.74\% vs. 69.34\%) and MCC (0.514 vs. 0.489). This indicates that abductive completion can sharpen the decision boundary when observed evidence alone leaves criterion atoms unresolved. 

\subsection{Complementarity Analysis}
\label{sec:intersection}

To evaluate whether CoT and our method are \emph{complementary} (rather than simply one dominating the other), we pair $\alpha$NeSy-CTM and CoT predictions on deterministically aligned shared samples and partition outcomes into four cells: \textit{both correct}, \textit{both wrong}, \textit{CoT-Correct}, and \textit{$\alpha$NeSy-CTM-Correct}. Figure~\ref{fig:intersection_stacked} visualizes this outcome distribution, with the detailed  Table in the Appendix.

Figure~\ref{fig:intersection_stacked} shows substantial shared agreement but also clear non-overlap between the two methods. In particular, $\alpha$NeSy-CTM contributes to 18\% of the paired cases, while CoT still contributes a smaller but meaningful set of unique wins.
Evidence of complementarity is present across the different datasets, but the dominant side changes with the task profile. In TREC 2021--2022, the two methods remain relatively balanced, whereas in TREC 2023, the advantage shifts more clearly toward $\alpha$NeSy-CTM.

\begin{figure}[t]
\centering 
\includegraphics[width=\columnwidth]{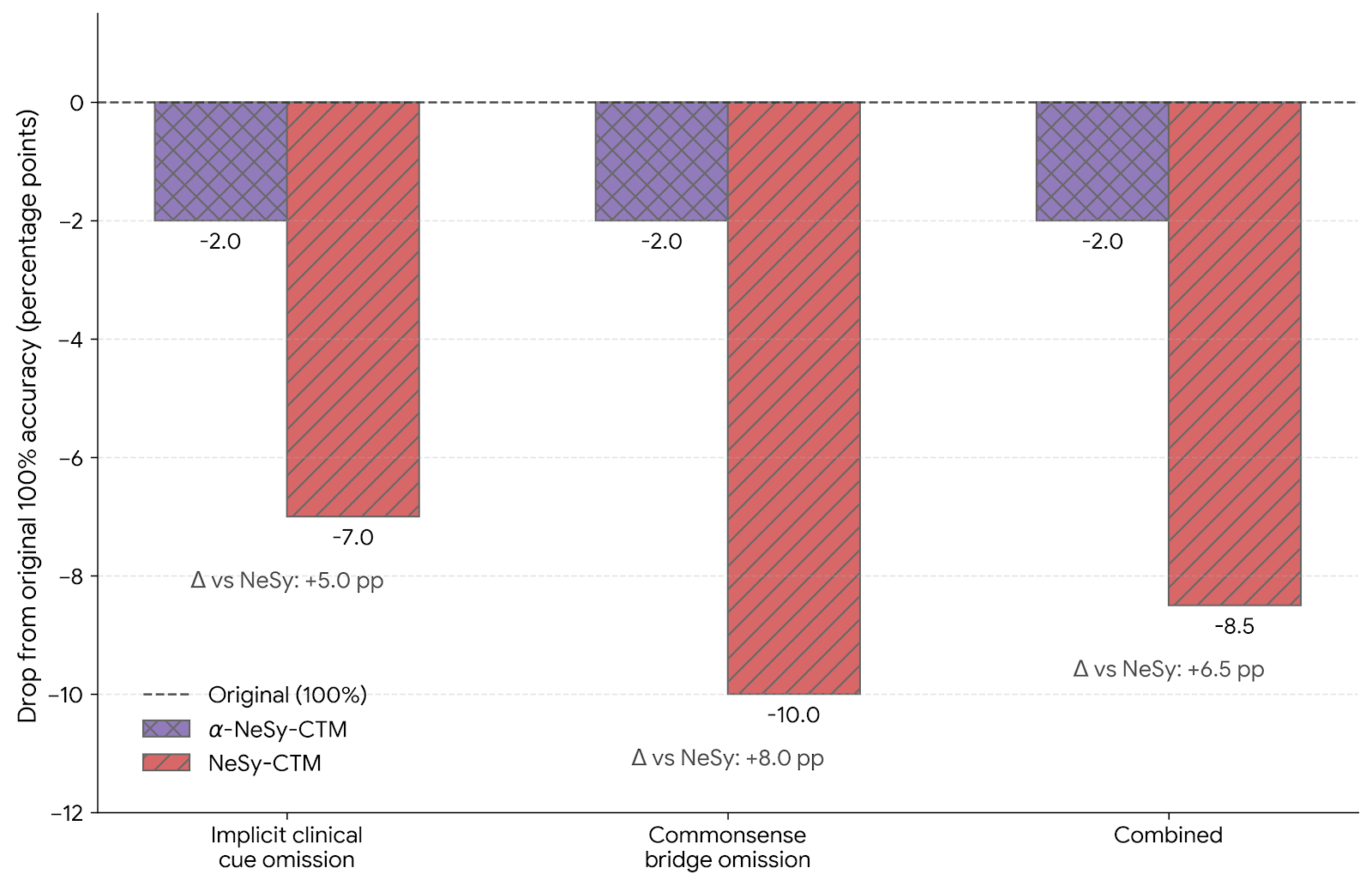}
\caption{Performance under controlled evidence perturbation. Bars show post-perturbation accuracy for NeSy-CTM and $\alpha$NeSy-CTM; the dashed line marks the original setting where both methods achieve 100\% accuracy before evidence masking. Labels inside each group show the percentage-point advantage of $\alpha$NeSy-CTM over NeSy-CTM.}
\label{fig:ab_levels}
\end{figure}

\subsection{Impact of Abductive Reasoning on Robustness}
\label{subsec:controlled-gap}

We evaluate the robustness of neurosymbolic CTM methods to missing evidence with a controlled perturbation study designed to preserve the original eligibility label. Starting from cases that both systems classify correctly, we automatically generate controlled perturbations using an LLM. For each case, the LLM is prompted to remove or weaken non-decisive evidence while strictly preserving the gold eligibility label. Details of perturbations and validation are in Appendix~\ref{app:perturbation_details}. 

We consider two gap types: \textit{implicit clinical cue omission}, where explicit mentions of criterion-relevant facts are removed but weaker contextual cues remain, and \textit{commonsense bridge omission}, where the link between an observed fact and an eligibility concept is weakened so that a short medical or commonsense inference is required. This setting tests whether abductive reasoning can use sparse but label-preserving context without making unsupported assumptions, and systematically investigates the impact of abductive reasoning on robustness in neurosymbolic CTM methods.

For each perturbed sample, we run $\alpha$NeSy-CTM and NeSy-CTM on the same input and compare predictions against the unchanged gold label.

\begin{figure}[t]
\centering
\includegraphics[width=\columnwidth]{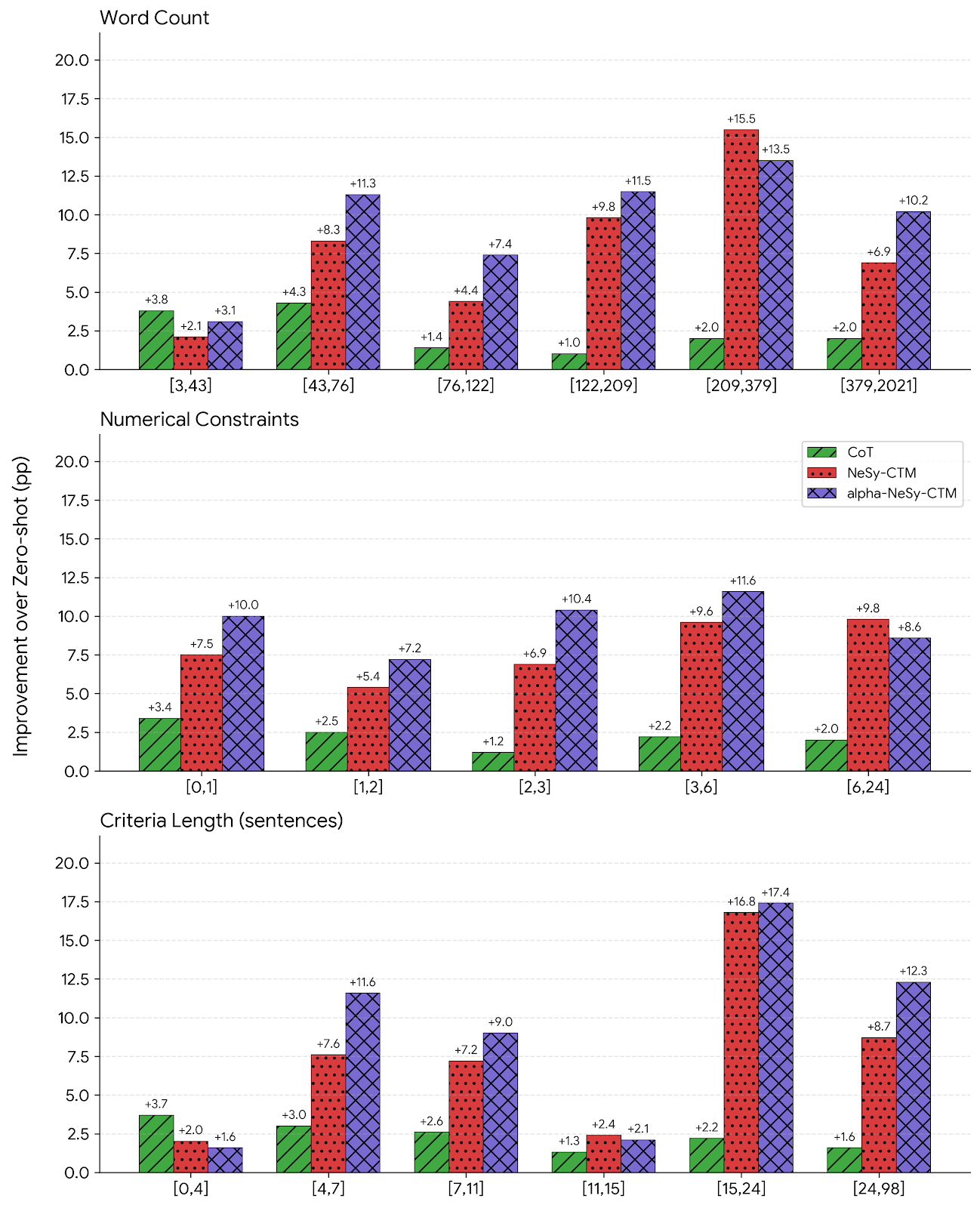}
\caption{Fine-grained complexity-bucket improvement over the Zero-shot baseline, averaged over the three fixed-$n$ backbones. Each panel reports one complexity dimension, and bars show percentage-point accuracy gains for CoT, NeSy-CTM, and $\alpha$NeSy-CTM relative to Zero-shot.}
\label{fig:complexity_analysis_full}
\end{figure}

Figure~\ref{fig:ab_levels} summarizes performance before and after controlled perturbation. The dashed reference line marks the original, unedited setting, where both models are fully correct by construction. After perturbation, the bars show that $\alpha$NeSy-CTM consistently outperforms NeSy-CTM across both omission types, achieving 98.00\% combined accuracy versus 91.50\%. Notably, the largest gain appears under commonsense bridge omission (+8.0 percentage points), suggesting that abductive reasoning is most useful when explicit evidence is weakened, requiring additional inference steps to make sparse contextual support explicit. 

Aggregated confusion counts show the same pattern: $\alpha$NeSy-CTM reduces false positives from 15 to 2 while keeping false negatives unchanged (2 in both systems). Thus, the benefit of abduction is not simply more permissive matching; it mainly preserves negative cases under incomplete evidence.

\subsection{Complexity Analysis}

To understand where the neurosymbolic gains come from, we perform a complexity-aware evaluation over the TREC 2021--2023 patient--trial pairs along three axes: word count, numerical-constraint density, and criteria sentence count. For readability, the main paper reports a coarse two-bin summary for each axis, while the original fine-grained bucketization is deferred to Appendix Table~\ref{tab:complexity_analysis_full}.

Figure~\ref{fig:complexity_analysis_full} is organized as three panels corresponding to word count, numerical-constraint density, and criteria sentence count. Within each bucket, bars report the improvement in 3-backbone average accuracy over the Zero-shot baseline, measured in percentage points, for CoT, NeSy-CTM, and $\alpha$NeSy-CTM. Zero-shot is therefore the zero line rather than a separate bar, making relative gains easier to compare across buckets.

The improvement chart highlights two patterns. First, neuro-symbolic methods generally deliver substantially larger positive gains over Zero-shot than standard CoT across most fine-grained buckets, a trend that becomes particularly pronounced as textual complexity escalates (e.g., word count > 43 or criteria length > 4 sentences) or when explicit numerical constraints are introduced. Second, this neuro-symbolic advantage exhibits distinct behaviors depending on the dimension of complexity: while CoT remains highly competitive and even outperforms symbolic integration in ultra-short text ranges (the first buckets of Word Count and Criteria Length), it lags significantly behind even under minimal numeric density (the [0,1) constraint bucket). This nuanced divergence underscores that the performance leaps of our framework are primarily catalyzed by the presence of explicit, formalizable decision structures (such as numerical thresholds and conjunctions).

\section{Related Work}
\label{sec:related}

Prior work on clinical trial matching and related clinical NLP falls into three broad lines. 

First, PLMs and LLMs, such as BioBERT, ClinicalBERT, and GPT-4, have been demonstrated to improve the semantic understanding of clinical text~\citep{lee2020biobert,huang2019clinicalbert}. 

Second, CTM systems have mainly framed the task as lexical, dense, or LLM-based matching, using methods such as BM25, ColBERT, semantically enriched patient--trial representations, TrialGPT, and zero-shot patient--trial matching with general LLMs~\citep{robertson2009probabilistic,khattab2020colbert,hassanzadeh2020matching,jin2024trialgpt,wornow2024zeroshotctm}. This line also includes task-specific benchmarks and surveys that stress multi-evidence inference, safety, and end-to-end recruitment workflows for clinical trials~\citep{jullien2023nli4ct,jullien2024safebiomedical,ghosh2025llmctr}. 

Third, neurosymbolic and solver-augmented approaches translate natural language into executable logic, symbolic intermediate forms, or verifier-guided reasoning traces for more faithful inference~\citep{xu-etal-2026-adaptive,garnelo2019reconciling,pan2023logic,xu2024symboliccot,quan2024explanationrefiner,ranaldi2025quasisymbolic,yuan2019criteria2query}. However, extensive studies investigating the impact of neurosymbolic approaches on CTM have yet to be undertaken. The closest related study was done by \citep{xu-etal-2026-adaptive}, which, however, while performing evaluation on CTM, mainly focuses on adaptive neurosymbolic methods across diverse tasks.

Related benchmarks, including TREC Precision Medicine and the TREC Clinical Trials track, established trial retrieval and matching as standardized evaluation settings~\citep{zhang2020trec,roberts2022overview,rybinski2023questionnaires}, while clinical-note benchmarks such as DiReCT highlight the gap between fluent generation and interpretable reasoning~\citep{wang2024direct}. 

Our method differs from these strands by targeting patient--trial matching with explicit SMT-based verification of numerical, temporal, and compositional eligibility constraints, while also introducing abductive completion for missing patient evidence. In this sense, it addresses the main gap left by pure neural methods, which are semantically flexible but non-deterministic on hard constraints, and pure rule-based approaches, which are verifiable but brittle under natural-language variability.

\section{Conclusion}
\label{sec:conclusion}

In this work, we demonstrated that bridging the representational gap between unstructured clinical narratives and formal eligibility logic is critical for the safe automation of CTM. Rather than relying solely on the semantic flexibility of LLMs---which frequently results in over-permissive matching and high false-positive rates---or the brittleness of purely symbolic rules, our abductive neurosymbolic framework ($\alpha$NeSy-CTM) establishes a more reliable operational paradigm.

Our findings highlight a shift in the sensitivity-specificity trade-off. By delegating deterministic constraint satisfaction to an SMT solver and utilizing abductive reasoning to conservatively infer missing clinical context, we substantially reduce the false-positive burden that typically plagues human-in-the-loop screening. Furthermore, the framework's robustness under controlled evidence perturbation confirms that abduction is not merely a mechanism for permissive matching, but a crucial component for preserving accurate negative classifications when evidence is sparse. 

Finally, our analysis reveals a complementarity between neurosymbolic verification and CoT reasoning, suggesting that future clinical AI systems will benefit from dynamic, hybrid routing policies. Ultimately, this research provides a viable, logically grounded blueprint for building auditable and trustworthy decision-support systems in high-stakes medical domains.

\section*{Limitations}

Despite advancing the foundations for LLM-driven neurosymbolic CTM, additional work is required before the proposed approach can be fully deployed in real-world settings. Limitations include:

\begin{enumerate}
    \item Computational overhead: End-to-end inference requires both LLM generation and symbolic solving, which may increase latency and cost for large-scale retrieval settings.
    \item Formalization brittleness: Performance still depends on accurate criterion parsing; small schema or operator errors can propagate into solver outcomes.
    \item Abductive risk control: Assumption generation can overfill missing facts in underspecified records, so human review and conservative policies are still necessary.
    \item External validity: Evaluation focuses on English TREC-style data; broader multilingual and real-world EHR validation is required before clinical deployment.
\end{enumerate}

\section*{Ethical Considerations}

This work studies clinical trial matching in a research setting rather than as a deployed clinical screening system. Although our experiments use publicly available data, such methods still raise important privacy, safety, and governance concerns. Any real-world deployment would require institutionally approved data access, strong de-identification procedures, access control, and audit logging because patient notes, derived facts, and symbolic traces can expose sensitive information.

There are also direct decision-support risks. False positives may surface trials for which a patient is actually ineligible, while false negatives may hide potentially beneficial options. These risks are especially salient in our abductive setting, where unsupported assumptions could lead to an overestimation of eligibility if not carefully constrained. For this reason, we view the system as a clinician-support tool rather than an autonomous enrollment system, and we recommend explicit abstention or human escalation when evidence is incomplete, contradictory, or ambiguous.

Finally, benchmark performance does not guarantee equitable behavior across populations, institutions, or documentation styles. Bias may arise from trial-language conventions, disease coverage, subgroup under-representation, or shifts between curated benchmarks and real clinical notes. Future work should therefore include subgroup auditing, calibration under distribution shift, and prospective review of how symbolic assumptions and eligibility decisions affect different patient groups.

\bibliographystyle{plainnat}
\bibliography{custom}

\appendix

\section*{Appendix}
The appendix is organized as follows. Appendix~\ref{app:dataset_construction} delineates the comprehensive protocols for data sourcing, and binary-task derivation, followed by detailed dataset statistics. Appendix~\ref{app:complexity_analysis_extra} tabulates the fine-grained categorizations of clinical criteria, corresponding to the complete results in Table~\ref{tab:complexity_analysis_full}. Appendix~\ref{app:perturbation_details} elucidates the end-to-end dataset perturbation pipeline and validation framework designed to evaluate the impact of abductive reasoning. Appendix~\ref{sec:appendix_supplementary_results} comprises supplementary diagnostic analyses, incorporating the full intersection matrix, the direct-inference diagnostic, and the criterion-type advantage profile. Appendix~\ref{sec:appendix_error_analysis} provides a systematic taxonomy of representative model errors. Finally, Appendix~\ref{sec:appendix_prompts} outlines the full suite of prompt templates, while Appendix~\ref{sec:appendix_z3} presents a worked example of the formal Z3 verification process.

\section{Dataset Statistics and Preprocess}
\label{app:dataset_construction}

We use the TREC Clinical Trials matching resources for 2021--2023. For each year, we use the topic file (patient descriptions), the qrels file (topic--trial relevance labels), and the corresponding clinical-trial XML corpus.

For each year, we construct topic--trial pairs by joining qrels with the corresponding topic descriptions.
Each instance contains \texttt{topic\_id}, \texttt{doc\_id}, \texttt{topic\_description}, and a relevance label in $\{0,1,2\}$.
Label $0$ corresponds to the non-relevant class in the original qrels, while labels $1$ and $2$ denote the two explicit decision classes retained in our preprocessing.

Our final evaluation is formulated as a binary classification task (ineligible vs. eligible) and uses only labels $1$ and $2$.
We exclude label $0$ because it does not provide a direct eligibility decision and would blur the main supervision signal when the objective is to evaluate whether a method can distinguish clearly ineligible cases from clearly eligible cases.
This binary formulation also matches the standard trial-screening setting, where the model must produce a concrete eligibility decision rather than an uncertain or background category.
For each year, we sample $100$ examples from label $1$ and $100$ examples from label $2$, resulting in a balanced 200-example evaluation set per year, details are shown in Table ~\ref{tab:app_dataset_stats}.
We select examples uniformly at random within each label and keep the same per-label quota across years to make the evaluation balanced and comparable.
Thus, all reported binary results are based on $600$ examples in total across 2021--2023.

\begin{table*}[t]
\centering
\small
\begin{tabular}{lrrrrrr}
\toprule
Year & Topics & Label $1$ Pool & Label $2$ Pool & Sampled $L1$ & Sampled $L2$ & Final Eval \\
\midrule
2021 & 75 & 6{,}019 & 5{,}570 & 100 & 100 & 200 \\
2022 & 50 & 3{,}036 & 3{,}939 & 100 & 100 & 200 \\
2023 & 37 & 10{,}699 & 11{,}667 & 100 & 100 & 200 \\
\midrule
Total & 162 & 19{,}754 & 21{,}176 & 300 & 300 & 600 \\
\bottomrule
\end{tabular}
\caption{Dataset statistics from preprocessing to final binary evaluation. "Label $1$ Pool" and "Label $2$ Pool" denote the available instances retained for the binary task after excluding label $0$; "Sampled $L1$" and "Sampled $L2$" denote the randomly selected evaluation examples (100 per label per year); "Final Eval" denotes the resulting balanced subset used for model evaluation.}
\label{tab:app_dataset_stats}
\end{table*}

\section{Complexity Analysis}
\label{app:complexity_analysis_extra}
Fine-grained complexity-bucket complexity analysis Table~\ref{tab:complexity_analysis_full}.

\begin{table*}[t]
\centering
\small
\setlength{\tabcolsep}{6pt}
\renewcommand{\arraystretch}{1.05}
\begin{tabular}{llccccc}
\toprule
Dimension & Bin & $N$ & $\alpha$NeSy-CTM (\%) & NeSy-CTM (\%) & CoT (\%) & Zero-shot (\%) \\
\midrule
\multirow{6}{*}{Word Count}
& [3,43]    & 96  & 69.1 & 68.1 & \textbf{69.8} & 66.0 \\
& [43,76]   & 100 & \textbf{72.0} & 69.0 & 65.0 & 60.7 \\
& [76,122]  & 100 & \textbf{67.7} & 64.7 & 61.7 & 60.3 \\
& [122,209] & 102 & \textbf{72.6} & 70.9 & 62.1 & 61.1 \\
& [209,379] & 101 & 77.2 & \textbf{79.2} & 65.7 & 63.7 \\
& [379,2021]& 101 & \textbf{73.6} & 70.3 & 65.4 & 63.4 \\
\midrule
\multirow{5}{*}{Numerical Constraints}
& [0,1]   & 136 & \textbf{71.8} & 69.3 & 65.2 & 61.8 \\
& [1,2]   & 129 & \textbf{70.3} & 68.5 & 65.6 & 63.1 \\
& [2,3]   & 87  & \textbf{75.9} & 72.4 & 66.7 & 65.5 \\
& [3,6]   & 132 & \textbf{73.5} & 71.5 & 64.1 & 61.9 \\
& [6,24]  & 116 & 69.8 & \textbf{71.0} & 63.2 & 61.2 \\
\midrule
\multirow{6}{*}{Criteria Length (sentences)}
& [0,4]   & 82  & 70.3 & 70.7 & \textbf{72.4} & 68.7 \\
& [4,7]   & 101 & \textbf{65.7} & 61.7 & 57.1 & 54.1 \\
& [7,11]  & 125 & \textbf{72.5} & 70.7 & 66.1 & 63.5 \\
& [11,15] & 82  & 72.4 & \textbf{72.7} & 71.6 & 70.3 \\
& [15,24] & 107 & \textbf{76.6} & 76.0 & 61.4 & 59.2 \\
& [24,98] & 103 & \textbf{74.1} & 70.5 & 63.4 & 61.8 \\
\midrule
\textbf{Overall} & - & \textbf{600} & \textbf{72.1} & 70.4 & 64.9 & 62.5 \\
\bottomrule
\end{tabular}
\caption{Fine-grained complexity-bucket performance averaged over the three fixed-$n$ backbones. This is the original bucketized version of Figure~\ref{tab:complexity_analysis_full}.}
\label{tab:complexity_analysis_full}
\end{table*}

\section{Perturbation Generation and Verification for the Impact of Abductive Reasoning}
\label{app:perturbation_details}

To ensure reproducibility and minimize human bias, all controlled perturbations in Section~\ref{subsec:controlled-gap} were generated automatically using a large language model (LLM), rather than by manual editing. This section details the end-to-end process, including model selection, prompt design, batch generation, and label-preserving quality control.
We used GPT-3.5-turbo (OpenAI API, 2024-03 version) to generate all text perturbations. All prompts and completions were processed via the official OpenAI API with temperature set to 0.2 for deterministic outputs. The same process is compatible with other strong LLMs (e.g., GPT-5, Qwen, or Claude) and can be reproduced with any model supporting system/user prompt roles.
For each original patient case, we constructed a prompt that (1) described the clinical context, (2) specified the gap type (implicit clinical cue omission or commonsense bridge omission), and (3) instructed the LLM to rewrite the patient profile by removing or weakening non-decisive evidence while strictly preserving the label-defining facts. The following template illustrates the prompt configuration for the implicit clinical cue omission gap type:
\begin{quote}
\small
\texttt{You are a clinical trial data editor. Given the following patient profile and eligibility criteria, rewrite the profile by hiding or weakening explicit mentions of facts that support the eligibility label, but keep enough indirect clues so that the original label remains unchanged. Do not add or remove any decisive evidence. Only edit non-decisive details.\
\
Patient profile: ...\
Eligibility criteria: ...\
Label: [eligible/ineligible]\
Gap type: [implicit clinical cue omission]\
Edited profile:}
\end{quote}
A similar template was used for the commonsense bridge omission gap type, instructing the model to abstract or weaken the bridge between observed facts and eligibility concepts.

All 100 original cases (50 TP, 50 TN) were processed in batches using a Python script. Each case was sent to the LLM with the appropriate prompt, and the returned edited profile was stored alongside the original for downstream evaluation. The process is fully automated and can be rerun with a single command. To ensure that perturbations did not inadvertently flip the gold label, we implemented a two-stage quality control:
\begin{itemize}
\item \textbf{Automatic check:} After LLM editing, the perturbed profiles were re-evaluated by the same eligibility classifier used in the main experiments. If the predicted label changed, the case was flagged for review.
\item \textbf{Manual review:} All flagged cases were inspected manually. If the LLM had removed or altered a label-defining fact, the edit was discarded and the prompt was adjusted for a new generation. 
\end{itemize}
This process is intended to ensure that perturbed samples remain label-preserving, combining automatic checks with manual inspection of flagged cases.

\section{Analyses and Detailed Supplementary Results}
\label{sec:appendix_supplementary_results}

\subsection{Full Intersection Matrix}

Appendix Table~\ref{tab:intersection} reports the exact counts and percentages underlying Figure~\ref{fig:intersection_stacked} in the main paper.

\begin{table*}[t]
\centering
\small
\setlength{\tabcolsep}{4.5pt}
\begin{tabular}{@{}lccccc@{}}
\toprule
\textbf{Year} & \textbf{N} & \textbf{Both $\checkmark$} & \textbf{Both $\times$} & \textbf{CoT-Correct} & \textbf{$\alpha$NeSy-CTM-Correct} \\
2021 & 200 & 112 (56.0\%) & 28 (14.0\%) & 31 (15.5\%) & 29 (14.5\%) \\
2022 & 200 & 123 (61.5\%) & 23 (11.5\%) & 24 (12.0\%) & 30 (15.0\%) \\
2023 & 200 & 105 (52.5\%) & 29 (14.5\%) & 19 (9.5\%) & 47 (23.5\%) \\
\midrule
Total & 600 & 340 (56.7\%) & 80 (13.3\%) & 74 (12.3\%) & 106 (17.7\%) \\
\bottomrule
\end{tabular}
\caption{Exact prediction intersection matrix underlying Figure~\ref{fig:intersection_stacked}. CoT-Correct: CoT correct and $\alpha$NeSy-CTM wrong; $\alpha$NeSy-CTM-Correct: vice versa.}
\label{tab:intersection}
\end{table*}

\subsection{Criterion-Type Advantage Profile}

For the manually audited disagreement subset discussed in Section~\ref{sec:intersection}, Appendix Table~\ref{tab:method_advantages} reports the criterion-type advantage profile.

\begin{table*}[t]
\centering
\small
\setlength{\tabcolsep}{6pt}
\renewcommand{\arraystretch}{1.10}
\begin{tabular}{@{}lcccc@{}}
\toprule
\multirow{2}{*}{\textbf{Criterion Type}} &
\multicolumn{2}{c}{\textbf{Tag Count}} &
\multirow{2}{*}{\textbf{CoT-Correct Share}} &
\multirow{2}{*}{\textbf{Method Edge}} \\
\cmidrule(lr){2-3}
& \textbf{$\alpha$NeSy-CTM-Correct} & \textbf{CoT-Correct} & & \\
\midrule
Numerical Constraint & 34 & 19 & 59.4\% & \textbf{$\alpha$NeSy-CTM (+15)} \\
Boundary Condition & 23 & 1 & 3.1\% & \textbf{$\alpha$NeSy-CTM (+22)} \\
Temporal Reasoning & 0 & 9 & 28.1\% & \textbf{CoT (+9)} \\
Multi-Condition Logic & 1 & 3 & 9.4\% & CoT (+2) \\
Negation / Exclusion & 1 & 0 & 0.0\% & \textbf{$\alpha$NeSy-CTM (+1)} \\
\midrule
\textbf{Combined} & \textbf{59} & \textbf{32} & \textbf{100.0\%} & --- \\
\bottomrule
\end{tabular}
\caption{Criterion-type advantage profile on disagreement cases.
\textit{Tag Count} reports criterion-type occurrences in a manually audited subset.
\textit{CoT-Correct Share} is normalized within the CoT-Correct tag pool (total $n{=}32$), indicating where CoT-exclusive recoveries concentrate.}
\label{tab:method_advantages}
\end{table*}

\section{Error Taxonomy Examples}
\label{sec:appendix_error_analysis}

Table~\ref{tab:appendix_error_taxonomy} is the Appendix~C companion of Section~\ref{sec:intersection}: it uses the same four taxonomy labels and maps each one to representative \texttt{(topic\_id, doc\_id)} examples.

\begin{table*}[!t]
\centering
\small
\setlength{\tabcolsep}{3pt}
\begin{tabular}{@{}p{2.1cm}p{2.1cm}p{2.8cm}p{4.1cm}p{3.8cm}@{}}
\toprule
\textbf{Error Type} & \textbf{Operational Definition} & \textbf{Concrete Case} & \textbf{Observed Failure Pattern} & \textbf{Recommended Fix} \\
\midrule
Missing Information / World Knowledge
& Key patient attributes are underspecified or absent, and the method fills gaps with weakly supported assumptions.
& \texttt{(20, NCT04679116)}
& In abductive reasoning, missing criteria (e.g., bilateral hernia, ASA class) are left unconstrained, allowing SAT models that overestimate eligibility; prediction becomes eligible while gold is ineligible.
& Constrain abduction with plausibility priors and ``minimal-assumption'' objective; require evidence tags for each hypothesized variable. \\

Solver / Constraint Error
& The solver pipeline fails to return a stable SAT/UNSAT decision under resource or refinement limits (neurosymbolic pipeline only).
& \texttt{(15, NCT00826163)}
& quasi-symbolic output returns label \texttt{0} with explanation \texttt{Max refinement rounds (3) reached. Could not generate valid JSON.}
& Add adaptive retry policy (depth and timeout scheduling), fallback parser, and calibrated abstention handling instead of hard failure. \\

Label Mapping / Decision Threshold Error
& Facts are mostly extracted, but final decision projection is wrong (e.g., inclusion/exclusion conflict resolution, threshold projection, or numerical-constraint application at label stage).
& \texttt{(65, NCT00127634)}
& CoT predicts eligible while exclusion evidence indicates ineligible; the failure occurs at output-stage label projection rather than sentence parsing.
& Add rule-grounded decision trace: enforce explicit ``inclusion satisfied / exclusion violated'' checklist, conflict-priority policy, and threshold sanity checks before final label. \\

Reading Comprehension / Coreference Error
& Narrative evidence is misread at clause level (scope, qualifier, negation scope, or referent), causing criterion-level semantic mismatch before final label projection.
& \texttt{(62, NCT03344276)} and \texttt{(33, NCT01525563)}
& Misreading stenosis qualification (symptomatic vs asymptomatic threshold) and missing extraction of the physician-intent clause lead to the wrong eligibility decision.
& Introduce criterion-targeted extractors: unit normalization, inequality parser, and clause-level NLI checks for mandatory fields. \\
\bottomrule
\end{tabular}
\caption{Appendix~C case mapping for the four error categories in Section~\ref{sec:intersection}.}
\label{tab:appendix_error_taxonomy}
\end{table*}

To make the failure modes auditable, Table~\ref{tab:appendix_error_taxonomy} reports representative examples for all four categories.
All case IDs are reported as \texttt{(topic\_id, doc\_id)}.

\section{Runtime Efficiency Analysis}
\begin{table*}[t]
\centering
\small
\caption{Runtime comparison across methods and models (seconds, mean $\pm$ std). All methods are evaluated on the same fixed 10 samples with 3 repeated no-cache runs per sample.}
\label{tab:runtime_main}
\begin{tabular}{lccc}
\toprule
Method & gpt-oss-120b & olmo-3-32b-think & qwen3-vl-32b-instruct \\
\midrule
Baseline (Zero-shot) & 1.50 $\pm$ 0.77 & 4.90 $\pm$ 4.04 & 6.29 $\pm$ 4.92 \\
Baseline (CoT) & 2.18 $\pm$ 0.88 & 12.87 $\pm$ 8.84 & 13.32 $\pm$ 5.45 \\
Quasi-Symbolic & 8.50 $\pm$ 2.95 & 62.58 $\pm$ 29.43 & 34.34 $\pm$ 12.94 \\
NeSy-CTM & 4.55 $\pm$ 1.37 & 77.86 $\pm$ 48.15 & 32.86 $\pm$ 38.13 \\
$\alpha$NeSy-CTM & 5.97 $\pm$ 2.87 & 81.17 $\pm$ 56.18 & 34.37 $\pm$ 29.10 \\
\bottomrule
\end{tabular}
\end{table*}

\noindent Table~\ref{tab:runtime_main} reports runtime under a fixed and reproducible protocol (same 10 samples, three no-cache runs per sample, identical method set across model backbones). Across all backbones, the ordering is stable: Baseline Zero-shot is fastest, Baseline CoT is second, and the neurosymbolic variants are slower. On gpt-oss-120b, relative to Zero-shot, mean latency increases by 3.03x for NeSy-CTM, 3.97x for $\alpha$NeSy-CTM, and 5.65x for Quasi-Symbolic. olmo-3-32b-think shows the largest absolute overhead (77.86-81.17s mean on LogicLM variants), with high variance (up to 56.18s std), qwen3-vl-32b-instruct exhibits a notable tail for NeSy-CTM (max 191.81s). Taken together, the results indicate a clear efficiency--reasoning trade-off: methods with stronger structured reasoning behavior require substantially higher inference-time budget.

All experiments in this study were conducted via API-based inference rather than local GPU deployment. The computational cost covers the runs included in the final manuscript, including the main experiments, ablation studies, and the limited pilot runs used to determine the final prompting and inference settings.

\begin{table*}[t]
\centering
\scriptsize
\caption{Detailed runtime statistics by model and method.}
\label{tab:runtime_detailed}
\begin{tabular}{llccccccc}
\toprule
Model & Method & $N_{ok}/N_{total}$ & Mean & Std & Median & P90 & Min & Max \\
\midrule
gpt-oss-120b & Baseline (Zero-shot) & 30/30 & 1.50 & 0.77 & 1.29 & 2.05 & 0.92 & 3.95 \\
gpt-oss-120b & Baseline (CoT) & 30/30 & 2.18 & 0.88 & 2.01 & 3.23 & 0.94 & 4.39 \\
gpt-oss-120b & QuaSAR & 30/30 & 8.50 & 2.95 & 8.25 & 11.15 & 4.59 & 18.89 \\
gpt-oss-120b & NeSy-CTM & 30/30 & 4.55 & 1.37 & 4.24 & 6.17 & 2.59 & 8.07 \\
gpt-oss-120b & $\alpha$NeSy-CTM & 30/30 & 5.97 & 2.87 & 5.60 & 7.43 & 2.94 & 16.16 \\

\midrule
olmo-3-32b-think & Baseline (Zero-shot) & 30/30 & 4.90 & 4.04 & 3.63 & 8.05 & 1.94 & 23.07 \\
olmo-3-32b-think & Baseline (CoT) & 30/30 & 12.87 & 8.84 & 11.05 & 24.53 & 4.48 & 47.23 \\
olmo-3-32b-think & QuaSAR & 30/30 & 62.58 & 29.43 & 53.42 & 98.63 & 31.09 & 138.40 \\
olmo-3-32b-think & NeSy-CTM & 30/30 & 77.86 & 48.15 & 72.64 & 132.37 & 10.33 & 215.25 \\
olmo-3-32b-think & $\alpha$NeSy-CTM & 30/30 & 81.17 & 56.18 & 73.54 & 156.60 & 12.72 & 209.13 \\
\midrule
qwen3-vl-32b-instruct & Baseline (Zero-shot) & 30/30 & 6.29 & 4.92 & 4.75 & 9.12 & 2.94 & 29.84 \\
qwen3-vl-32b-instruct & Baseline (CoT) & 30/30 & 13.32 & 5.45 & 12.23 & 19.65 & 5.89 & 24.44 \\
qwen3-vl-32b-instruct & QuaSAR & 30/30 & 34.34 & 12.94 & 31.69 & 51.07 & 18.00 & 67.70 \\
qwen3-vl-32b-instruct & NeSy-CTM & 30/30 & 32.86 & 38.13 & 16.35 & 65.41 & 5.11 & 191.81 \\
qwen3-vl-32b-instruct & $\alpha$NeSy-CTM & 30/30 & 34.37 & 29.10 & 24.54 & 74.96 & 7.90 & 110.91 \\
\midrule
\bottomrule
\end{tabular}
\end{table*}

\section{Prompt Templates}
\label{sec:appendix_prompts}
This section reports the exact prompt templates used in our pipelines. 

\subsection{Template A: $\alpha$NeSy-CTM Prompt}
\begin{lstlisting}
"""
Unified Step: Perform abductive reasoning and formalize into Z3 code in one go.
"""
system = "You are a Z3 formalization expert. \n" \
         "Output Format:\n1. Abductive Reasoning (Briefly explain logic)\n2. Z3 Code (Wrapped in a ```z3 block)"
user = f"""Patient: {patient}
Trial: {trial}
Previous error (fix it): {feedback}

Generate Z3 code:
1.  **Internal Reasoning (Abduction)**:
    - First, analyze the patient data to infer implicit facts. For example, if "patient is an adult", you should use `age >= 18` in the Z3 code. If a patient has "metastatic cancer", you should use `is_metastatic = True`.
    - Then, compare the patient data against the trial criteria to identify missing information.
2.  **Define Variables**: Based on your reasoning, represent all relevant attributes as Z3 variables (Int, Bool, Real).
3.  **Handle Missing Information**: For any information identified as missing (e.g., 'ECOG_status'), create a Z3 variable for it (e.g., `ECOG_status = Int('ECOG_status')`) but do not add a fact to constrain its value. This allows the solver to explore possibilities.
4.  **Formulate Logic**:
    1. **ASCII ONLY**: STRICTLY use only standard ASCII characters for all operators. NEVER use Unicode symbols like >=, <=, ~, !=, etc. Use (>= X Y), (<= X Y), and (!= X Y).
    2. **SMT-LIB FORMAT**: Ensure all function calls (e.g., And, Or, define-fun, assert) use correct SMT-LIB syntax.
        * **Parameter Separation**: do not forget to use commas where needed.
        * **Parentheses**: Every function call MUST be fully enclosed in parentheses, including nested calls.
    3. **Approximation**: Represent "approximately equal" (\~) using an epsilon range: |X-Y| < Epsilon.

Here is an example to follow:
==== Starting SMT Formalisation ====   
Input Premise:    
Patient 001.

A 52-year-old male with long-standing type 2 diabetes, essential hypertension, and anaemia of chronic disease presents as a former smoker who quit a decade ago; he has completed his primary COVID-19 vaccination series, is not on immuno-suppressive therapy, takes a stable ACE-inhibitor, and is not taking an SGLT2 inhibitor, bisphosphonate, or opioids. He has no history of malignancy, COPD, migraine, heart failure, or autoimmune disease, and owns a smartphone suitable for digital interventions. Recent measurements show haemoglobin 11.2 g/dL, creatinine clearance 72 mL/min, platelet count 110 x 10^3 cells/uL, eGFR 60 mL/min/1.73 m^2, BMI 28 kg/m^2, systolic blood pressure 148 mmHg, HbA1c 7.5 %, and fasting glucose 115 mg/dL, while other parameters (WBC, ANC, FEV1%, bone-density T-score, LVEF, monthly headache days, PHQ-9 score) have not yet been collected.

Trial A - Chronic-Anaemia Therapy.

Adults aged 18-65 who suffer from anaemia of chronic disease may enter this study if their latest labs show haemoglobin at least 9 g/dL, a creatinine-clearance of 60 mL/min or better, and a platelet count of 100 x 10^3 cells/uL or higher; participants must not be pregnant and must have no active malignancy.


LLM Response (abductive) :

First step - Abductive inference:

Add implications and a list of missing facts.


To convert the given natural language description into SMT-LIB format, we need to identify the relevant variables, their types, and the constraints that determine whether Patient 001 is eligible for Trial A.

1. **Variables and Types:**

- Age: Int

- Gender: Bool (true for male, false for female)

- Haemoglobin: Real

- CreatinineClearance: Real

- PlateletCount: Real

- PregnancyStatus: Bool (true for pregnant, false for not pregnant)

- ActiveMalignancy: Bool (true for active malignancy, false for no active malignancy)

2. **Constraints:**

- Age must be between 18 and 65.

- Haemoglobin must be at least 9 g/dL.

- Creatinine clearance must be at least 60 mL/min.

- Platelet count must be at least 100 x 10^3 cells/uL.

- Patient must not be pregnant.

- Patient must have no active malignancy.

==== SMT Formalisation Complete ====

Initial code:

; Declare variables and their types

(declare-const Age Int)

(declare-const Gender Bool) ; true for male, false for female

(declare-const Haemoglobin Real) ; in g/dL

(declare-const CreatinineClearance Real) ; in mL/min

(declare-const PlateletCount Real) ; in 10^3 cells/uL

(declare-const PregnancyStatus Bool) ; true for pregnant, false for not pregnant

(declare-const ActiveMalignancy Bool) ; true for active malignancy, false for no active malignancy

; Assert the known values for Patient 001

(assert (= Age 52))

(assert (= Gender true)) ; Male

(assert (= Haemoglobin 11.2))

(assert (= CreatinineClearance 72))

(assert (= PlateletCount 110))

(assert (= PregnancyStatus false)) ; Not pregnant

(assert (= ActiveMalignancy false)) ; No active malignancy

; Assert the eligibility constraints for Trial A

(assert (and (>= Age 18) (<= Age 65))) ; Age constraint

(assert (>= Haemoglobin 9)) ; Haemoglobin constraint

(assert (>= CreatinineClearance 60)) ; Creatinine clearance constraint

(assert (>= PlateletCount 100)) ; Platelet count constraint

(assert (not PregnancyStatus)) ; Must not be pregnant

(assert (not ActiveMalignancy)) ; Must have no active malignancy

; Check satisfiability

(check-sat)
    
Only code, no markdown.
\end{lstlisting}

\subsection{Template B: NeSy-CTM Prompt}
\begin{lstlisting}
system = "You are a Z3 formalization expert. Output ONLY executable Z3 code. You MUST wrap the entire code in a single '```z3' code block."
user = f"""Patient: {patient}
Trial: {trial}
Previous error (fix it): {feedback}

Generate Z3 code:
1.  **Define Variables**: Represent all patient and trial attributes as Z3 variables (Int, Bool, Real).
2.  **Handle Missing Information**: If a criterion requires information not present in the patient data (e.g., 'EGFR_status'), create a Z3 variable for it (e.g., `EGFR_status = Bool('EGFR_status')`) but do not add a fact to constrain its value.
3.  **Formulate Logic**:
    1. **ASCII ONLY**: STRICTLY use only standard ASCII characters for all operators. NEVER use Unicode symbols like >=, <=, ~, !=, etc. Use (>= X Y), (<= X Y), and (!= X Y).
    2. **SMT-LIB FORMAT**: Ensure all function calls (e.g., And, Or, define-fun, assert) use correct SMT-LIB syntax.
        * **Parameter Separation**: do not forget to use commas where needed.
        * **Parentheses**: Every function call MUST be fully enclosed in parentheses, including nested calls.
    3. **Approximation**: Represent "approximately equal" (~) using an epsilon range: |X-Y| < Epsilon.

Here is an example to follow:
==== Starting SMT Formalisation ====
    Input Premise:
Patient 001.

A 52-year-old male with long-standing type 2 diabetes, essential hypertension, and anaemia of chronic disease presents as a former smoker who quit a decade ago; he has completed his primary COVID-19 vaccination series, is not on immuno-suppressive therapy, takes a stable ACE-inhibitor, and is not taking an SGLT2 inhibitor, bisphosphonate, or opioids. He has no history of malignancy, COPD, migraine, heart failure, or autoimmune disease, and owns a smartphone suitable for digital interventions. Recent measurements show haemoglobin 11.2 g/dL, creatinine clearance 72 mL/min, platelet count 110 x 10^3 cells/uL, eGFR 60 mL/min/1.73 m^2, BMI 28 kg/m^2, systolic blood pressure 148 mmHg, HbA1c 7.5 %, and fasting glucose 115 mg/dL, while other parameters (WBC, ANC, FEV1 %, bone-density T-score, LVEF, monthly headache days, PHQ-9 score) have not yet been collected.

Trial A - Chronic-Anaemia Therapy.
### Z3 Python Code Output:

```z3
s.add(patient_facts)
s.add(eligibility_criteria)

- Age: Int

- Haemoglobin: Real
3.  **Trial Criteria**: Create a list `eligibility_criteria` for all trial requirements.
4.  **Solver Logic**: Create a `Solver` instance `s`, add both lists to it.
5.  **Set Result**: Use an `if s.check() == sat:` block to set a variable `result` to "2" (Eligible) or "1" (Ineligible).


- PregnancyStatus: Bool (true for pregnant, false for not pregnant)
2. **Constraints:**

- Age must be between 18 and 65.


- Platelet count must be at least 100 x 10^3 cells/uL.

- Patient must not be pregnant.

- Patient must have no active malignancy.

Here is the SMT-LIB model:

This model checks whether Patient 001 satisfies all the eligibility criteria for Trial A based on the given constraints.

==== SMT Formalisation Complete ====
Initial code:

; Declare variables and their types

(declare-const Age Int)

(declare-const Gender Bool) ; true for male, false for female

(declare-const Haemoglobin Real) ; in g/dL

(declare-const CreatinineClearance Real) ; in mL/min

(declare-const PlateletCount Real) ; in 10^3 cells/uL

(declare-const PregnancyStatus Bool) ; true for pregnant, false for not pregnant

(declare-const ActiveMalignancy Bool) ; true for active malignancy, false for no active malignancy

; Assert the known values for Patient 001

(assert (= Age 52))

(assert (= Gender true)) ; Male

(assert (= Haemoglobin 11.2))

(assert (= CreatinineClearance 72))

(assert (= PlateletCount 110))

(assert (= PregnancyStatus false)) ; Not pregnant

(assert (= ActiveMalignancy false)) ; No active malignancy

; Assert the eligibility constraints for Trial A

(assert (and (>= Age 18) (<= Age 65))) ; Age constraint

(assert (>= Haemoglobin 9)) ; Haemoglobin constraint

(assert (>= CreatinineClearance 60)) ; Creatinine clearance constraint

(assert (>= PlateletCount 100)) ; Platelet count constraint

(assert (not PregnancyStatus)) ; Must not be pregnant

(assert (not ActiveMalignancy)) ; Must have no active malignancy

; Check satisfiability

(check-sat)
Only code, no markdown."""
\end{lstlisting}

\subsection{Template C: QuaSAR Prompt}
\begin{lstlisting}
"""Step 1: Use LLM to formalize natural language into a Z3 code string."""
prompt = f"""You are a Z3 formalization expert. Your task is to convert the following patient and trial information into a Z3 Python code representation.
Do not solve it. Just provide the code that represents the problem.

# Patient Data
{patient_data}

# Trial Criteria
{trial_criteria}

Generate the Z3 code representation. Output ONLY the Python code block.
"""
"""Step 2: Use LLM to reason over the formalized code and provide a final answer."""
prompt = f"""You are an experienced expert skilled in answering complex problems through logical reasoning.

#Task
Analyze the following Z3 code, which represents a patient's eligibility for a clinical trial. Determine the final eligibility and provide a step-by-step explanation.

#Z3 Code
```python
{formalized_code}
```

#Steps
1. **Analysis**: Explain what the patient facts and trial rules are based on the Z3 code.
2. **Reasoning**: Logically deduce whether the patient is Eligible, Ineligible.
3. **Answering**: Conclude with the final answer in the format: <decision>LABEL</decision>, where LABEL is one of 'Eligible', 'Ineligible'.

#Previous Error (if any, please fix your reasoning)
{feedback}
"""
\end{lstlisting}

\subsection{Template D: Chain-of-Thought Prompt}
\begin{lstlisting}
f"Patient description:\n{topic_description}\n\n"
f"Clinical trial document:\n{doc_content}\n\n"
"Task: Determine the patient's eligibility for the clinical trial based on the provided information.\n"
"Answer Options: (1) 1(ineligible), (2) 2 (eligible)\n"
"Instructions: As a clinical trial screening expert, follow these steps to determine eligibility:\n"
"Reasoning Steps: 1.Identify Criteria: Summarize the primary inclusion and exclusion criteria (relevant to the patient description) found in the Clinical Trial Document. 2.Evaluate Patient: Compare the Patient Description against the identified criteria. State explicitly if the patient meets inclusion criteria or violates any exclusion criteria. 3.Conclusion: Based on the comparison, conclude whether the patient is eligible or ineligible. Select the single best option from the list (1 or 2) that matches your determination." 
"Output Format: Make sure start with your final answer(it should be the Predicted_label, not other numbers):\n, 1 for ineligible, 2 for eligible; followed by your thought process step-by-step"
\end{lstlisting}

\subsection{Template E: Zero-Shot Prompt}
\begin{lstlisting}
f"Patient description:\n{topic_description}\n\n"
f"Clinical trial document:\n{doc_content}\n\n"
"Question: Based on the patient description and clinical trial document, determine the patient's eligibility for the clinical trial.\n"
"Answer Choices: (1) 1 (ineligible) (2) 2 (eligible)\n"
"Instructions: This is a zero-shot classification task. Select the single best option that matches your determination.\n"
"Output Format: Make sure start with your final answer(it should be thePredicted_label, not other numbers):\n, 1 for ineligible, 2 for eligible; followed by the reason for your decision."
\end{lstlisting}

\section{Worked Z3 Verification Example}
\label{sec:appendix_z3}

We provide a compact end-to-end example showing how trial constraints and patient facts are verified in Z3. We keep the code in \texttt{verbatim} so that the SMT-LIB syntax, indentation, and inline comments remain exactly readable to reviewers. The first block encodes the trial rule once; the next two blocks instantiate two patients with \texttt{push}/\texttt{pop} so that the same base constraints can be reused without rewriting the trial logic.

\subsection{Step 1: Trial Constraint Encoding}

\begin{lstlisting}
; Step 1: Encode the trial-level eligibility rule
; Eligible iff age is within range, HbA1c is below threshold,
; and there is no prior chemotherapy history.

(set-logic ALL)

; Declare variables used by both the trial rule and patient facts
(declare-const Age Int)
(declare-const HbA1c Real)
(declare-const Prior_Chemo Bool)

; Add inclusion/exclusion constraints derived from the trial text
(assert (>= Age 18))
(assert (<= Age 65))
(assert (< HbA1c 7.0))
(assert (not Prior_Chemo))
\end{lstlisting}

\subsection{Step 2: Verify Patient A (Expected SAT)}

\begin{lstlisting}
; Step 2: Reuse the same trial rule for Patient A
; push/pop keeps the base trial constraints unchanged for later cases.
(push)
(assert (= Age 45))
(assert (= HbA1c 6.8))
(assert (= Prior_Chemo false))
; Check whether the patient facts satisfy all trial constraints
(check-sat)
(get-model)
(pop)
\end{lstlisting}

\textbf{Z3 output (Patient A):}
\begin{verbatim}
sat
(model
    (define-fun Age () Int 45)
    (define-fun HbA1c () Real (/ 34 5))
    (define-fun Prior_Chemo () Bool false)
)
\end{verbatim}

\textbf{Interpretation:} All constraints are satisfied, so the patient is logically eligible.

\subsection{Step 3: Verify Patient B (Expected UNSAT)}

\begin{lstlisting}
; Step 3: Reuse the same trial rule for Patient B
; The prior-chemotherapy fact should violate the exclusion constraint.
(push)
(assert (= Age 45))
(assert (= HbA1c 6.8))
(assert (= Prior_Chemo true))
; This time Z3 should detect a logical conflict
(check-sat)
(pop)
\end{lstlisting}

\textbf{Z3 output (Patient B):}
\begin{verbatim}
unsat
\end{verbatim}

\textbf{Interpretation:} Patient B violates \texttt{(not Prior\_Chemo)} and is excluded, even though age and HbA1c satisfy thresholds.

\section{Artifact Availability, Licensing, and AI Usage}

The implementation code, prompt templates, evaluation scripts, and supplementary resources used in this work are provided through the anonymous repository link included in the first-page footnote for reviewing purposes.

This work uses publicly available benchmark resources from the TREC Clinical Trials tracks (2021--2023). The datasets are used in accordance with their original distribution terms and are not redistributed as part of this submission.

The experiments additionally rely on external software and model APIs, including the Z3 SMT solver and OpenAI API services for controlled perturbation generation. Open-source model backbones (e.g., Qwen and OLMo variants) are used under their respective licenses and usage terms.

All generated perturbations were used solely for research and evaluation purposes within the controlled experimental setting described in this paper. Large language models were used as experimental components.  We acknowledge the use of AI assistants to enhance writing clarity.

\end{document}